\documentclass{article}
\usepackage{spconf,amsmath,graphicx,amssymb, booktabs}
\usepackage{multirow}
\usepackage{hyperref}

\usepackage{url}
\usepackage{hyperref}
\hypersetup{colorlinks=True, urlcolor=black}

\newcommand{\Lm}{\mathcal{L}}
\newcommand{\Wm}{\mathcal{W}}

\newcommand{\myparagraph}[1]{\vspace{.4em} \noindent \textbf{#1}\ }
\usepackage{xcolor}
\usepackage{tipa}
\title{JOIST: A Joint Speech and Text Streaming Model For ASR}
\name{
\begin{tabular}{c} Tara N. Sainath, Rohit Prabhavalkar, Ankur Bapna, Yu Zhang, \\ Zhouyuan Huo, Zhehuai Chen, Bo Li, Weiran Wang, Trevor Strohman \end{tabular}
}
\address{Google, Inc. \\
\fontsize{9}{9}\selectfont\ttfamily\upshape
\{tsainath, prabhavalkar\}@google.com}
\begin{document}
\ninept
\maketitle
\begin{abstract}

\end{abstract}
We present JOIST, an algorithm to train a streaming, cascaded, encoder end-to-end (E2E) model with both speech-text paired inputs, and text-only unpaired inputs. Unlike previous works, we explore joint training with both modalities, rather than pre-training and fine-tuning. In addition, we explore JOIST using a streaming E2E model with an order of magnitude more data, which are also novelties compared to previous works. Through a series of ablation studies, we explore different types of text modeling, including how to model the length of the text sequence and the appropriate text sub-word unit representation. We find that best text representation for JOIST improves WER across a variety of search and rare-word test sets by 4--14\% relative, compared to a model not trained with text. In addition, we quantitatively show that JOIST maintains streaming capabilities, which is important for good user-level experience.

\section{Introduction}
Research on E2E automatic speech recognition (ASR) systems has become increasingly prominent in recent years, with multiple research groups demonstrating the strong performance of these models~\cite{li2020comparison, Ryan19, CC18, KimHoriWatanabe17, JinyuLi2019, Zeyer2020}.
However, training such E2E models comes with its own set of challenges -- most notably, E2E models are trained using large amounts of audio-text pairs; obtaining such hand-transcribed data is expensive and insufficient to fully cover the space of all possible words that might need to be recognized.
Thus, E2E models tend to perform poorly on utterances containing words that appear infrequently in the training data (e.g., named entities)~\cite{sainath2021cascadedlm}. 

A common solution to this issue is to leverage an external neural language model (LM), trained on a much larger amount of unpaired text data, which can be incorporated into an E2E model during decoding: e.g., shallow fusion~\cite{Chorowski17, Anjuli18}, or rescoring~\cite{sainath2021cascadedlm}. 
While such techniques can improve rare word recognition, they have limitations: for resource-constrained tasks such as on-device speech recognition, the additional memory required to store the LM might be prohibitive (e.g. 128M parameters for the LM, relative to $\sim$150M parameters for the base E2E model in~\cite{sainath2021cascadedlm}); the high cost of running an LM at each step of the beam search may necessitate second-pass rescoring instead of shallow fusion, thus limiting the scope for improvement since with rescoring the model can only correctly recognize words which are present in the first-pass decoded lattice.

An alternative approach to address the rare-word issue is to directly train the E2E model with unpaired text-only data.
Some of the earliest works in this direction have examined using text-to-speech systems to convert the text into paired audio-text pairs~\cite{ChenZhangRosenberg21}, or by incorporating cycle consistency losses (i.e., combining text-to-speech and speech-to-text losses)~\cite{Hori19, TjandraSaktiNakamura17}.
In addition, there has been work on injecting text into attention-based encoder-decoder models~\cite{YusufGandheSokolov22,Sainath2020b}.
An alternative approach focuses instead on creating a shared embedding space for the two modalities -- speech and text~\cite{Ankur2021, Ankur2022, Tang2022, Thomas2022, Zhehuai2022, ChungZhuZeng21, AoWangZhou22} -- thus improving ASR performance without increasing model parameters or decoding complexity. 
The use of a shared space allows training the E2E model with losses derived from paired or unpaired data, as discussed in Section~\ref{sec:related_work}.
In the present work, we build upon these modality matching techniques in order to make them suitable for large-scale streaming state-of-the-art cascaded recurrent neural transducer (RNN-T) models~\cite{arun21cascade, Sainath2022}. The contributions of this work include the following: (1) Unlike many previous works which have been applied to full-context models (i.e., full-utterance processing), we focus on streaming models~\cite{Ryan19} where words must be emitted as quickly as possible after the user speaks. We explore text injection as part of the streaming cascaded encoder model~\cite{arun21cascade}, which only has an acoustic look-ahead of 900ms, thus allowing for efficient low-latency streaming decoding. (2) Given the large scale of data available for our task, we explore joint training with a combination of losses computed on the supervised audio-text pairs along with the unpaired text data. We are guided by the intuition that pre-training followed by fine-tuning could be prone to forgetting the pre-trained task given the large amounts of supervised data~\cite{Junwen2022}. For example, past-research has shown that pre-training has a larger impact for tasks when supervised training data is limited (less then 1,000 hours)~\cite{Abdo2010,Sainath2011,Seide2011} compared to large data sets with tens of thousands of hours and medium-sized models~\cite{Sak14}. In addition, joint-training allows for a simpler training procedure, which is important with large-scale datasets. (3) We develop a simple solution to inject text that avoids the need for a sophisticated but more complex duration model to accurately model expected token durations. (4) Finally, we additionally optimize the model for ASR specific performance using the minimum word error rate criterion (MWER)~\cite{prabhavalkar2018minimum} on unpaired text data by modifying the standard formulation that is only applied to audio-text pairs.
  
We perform a series of ablation studies on a large vocabulary voice search task to understand the performance of different text-injection schemes, both with respect to how to model duration and what type of subword unit (i.e., word-pieces or phonemes), should be used to represent the unpaired text.
We find that proposed streaming JOIST configuration offers between a 4--14\% relative improvement in WER across a variety of voice search and rare-word sets, compared to a baseline system that does not use any unpaired text.

\section{Related Work} \label{sec:related_work}
The basic paradigm for training E2E models requires transcribed audio-text pairs; model performance is thus limited by the amount of training data~\cite{RohitSeq17, IriePrabhavalkarKannan19, Arun19}.
Many recent works have focused on improving E2E models by leveraging unpaired text, speech, or both.

Previous works which have investigated the use of unpaired speech data have focused on contrastive (e.g.,~\cite{VanDenOordLiVinyals18, SchneiderBaevskiCollobert19}) or reconstruction (e.g.,~\cite{ChungGlass20, LiuYangChi20}) losses.
Self-supervised learning approaches in natural language processsing (NLP), e.g., BERT~\cite{Devlin2018}, use masked language modeling (MLM) losses to pre-train encoders for NLP tasks -- this leverages the fact that NLP uses discrete input representations unlike speech.
Researchers have adapted these techniques for speech by deriving discrete labels for speech frames: e.g., using nearest neighbors (HuBERT~\cite{HsuBolteTsai21}); vector-quantization through Gumbel softmax or online K-means (vq-wav2vec~\cite{BaevskiSchneiderAuli20}) or a random (but deterministic) quantizer~\cite{ChiuQinZhang22}.
The wav2vec 2.0 system~\cite{BaevskiZhouMohamed20} proposed to combine the two steps -- quantization and contrastive losses -- which were subsequently combined with MLM losses in w2v-BERT~\cite{ChungZhangHan21}. 
All of these works have adopted the procedure of pre-training models with unpaired speech followed by fine-tuning the models on the paired audio-text data; notable exceptions include~\cite{Junwen2022, TalnikarLikhomanenkoCollobert21}, which jointly train on both losses in a multi-task framework.
Such techniques tend to achieve large gains when the amount of paired speech is limited~\cite{BaevskiZhouMohamed20, ChungZhangHan21}, for example on Librispeech~\cite{PanayotovChenPovey15}.
However, in previous work it has been observed that the gains from these techniques are limited when training with large-scale datasets~\cite{Sainath2020b}.
Therefore, in this work, we focus on techniques which incorporate unpaired text data into the E2E model.

There have also been recent works investigating the use of unpaired text data into the E2E model. 
\cite{ChenZhangRosenberg21} converts the unpaired text data into audio utterances using a text-to-speech (TTS) system, thus making the data amenable to supervised training.
A related approach consists of using cycle consistency losses -- using TTS in combination with ASR to train with unpaired data~\cite{Hori19, TjandraSaktiNakamura17}. 
One of the main disadvantages of these techniques is the high computational cost involved in converting text into audio through TTS. Another approach is to distill knowledge from an LM into the E2E model \cite{Yotaro2022}. 

An alternative approach, most closely related to our work, focuses on mapping the two modalities, audio and text, into a shared space.
For example, Bapna et al. propose SLAM which uses MLM losses for the text and w2v-BERT losses for the unpaired speech to learn audio/text representations, with additional losses to align the two modalities~\cite{Ankur2021}; the work is further generalized in the mSLAM approach to use multiple languages.
A similar approach -- dubbed STPT by Tang et al.~\cite{Tang2022} -- uses BART~\cite{LewisLiuGoyal19} and wav2vec 2.0~\cite{BaevskiZhouMohamed20} to train on unpaired text and speech, respectively, along with phoneme prediction and standard ASR losses on the paired data to align representations; SPLAT, proposed by Chung et al.~\cite{ChungZhuZeng21}, uses masked reconstruction losses for unpaired speech, a pre-trained BERT model for the unpaired text and a set of alignment losses (token-level or sequence-level) to align representations. 
The SpeechT5 system of Ao et al.~\cite{AoWangZhou22} utilizes an encoder-decoder model which operates in a shared latent space; the system is combined with a set of pre-nets (to map speech/text into the shared encoder/decoder input representation) and post-nets which map the decoder output into speech/text.
All of the above mentioned works focus on encoder-decoder architectures which are non-streaming; in this work, we focus on recognition using streaming RNN-T models~\cite{GravesMohamedHinton13, Variani20}.

Our work is most closely related to two recent works that have investigated techniques to incorporate text-only data into RNN-T based models~\cite{Thomas2022, Zhehuai2022}.
Thomas et al.~\cite{Thomas2022} propose a \emph{textogram} -- an input representation created by repeating one-hot embeddings of each input text tokens a fixed number of times (graphemes, in~\cite{Thomas2022}).
The textograms are stacked together with standard log-mel features along the time dimension.
When training on text-only data, the input log-mel features are set to zero; when training on audio-text paired data, the textogram features are set to zero; in either case, the model is trained with the standard RNN-T loss on the output text tokens. 
In order to ensure that the task of mapping from the textogram to output text tokens is not trivial, a subset of the input textogram features are masked.
The model, once pre-trained, is fine-tuned for a downstream spoken language understanding task.
In MAESTRO, Chen et al.~\cite{Zhehuai2022} propose to up-sample the input text tokens using a duration prediction model similar to that used in TTS, which is jointly trained with the rest of the model. 
In order to ensure that the representations learned from speech and text are aligned, the MAESTRO approach adds consistency losses to align the output representations from the two modalities using the paired data.
Although both of the aforementioned works~\cite{Thomas2022, Zhehuai2022} are applied to  RNN-T models, these works use full-context encoders (i.e., full utterance processing) and have not been explored in the context of streaming ASR.

\section{JOIST: Improving E2E ASR with Unpaired Text} \label{sec:joist}

In the present work, we simplify and expand on the techniques presented in~\cite{Thomas2022, Zhehuai2022} to build a solution for streaming RNN-T models~\cite{arun21cascade, Sainath2022}.
We assume that we have examples of transcribed audio-text pairs: $\mathcal{S}=\{(x_s, y_s)\}$ (in this work, $x_s$ corresponds to stacked log-mel feature frames; $y_s$ corresponds to word-pieces~\cite{Schuster2012}).
In addition, we assume that we have (a much larger) set of unpaired text data, $\mathcal{T}$.
Since the text data can be tokenized in multiple ways (e.g., as a sequence of phonemes, or word-pieces), for notational convenience we also represent the unpaired text data as a pair: $\mathcal{T} = \{(x_t, y_t)\}$ (in this work, $x_t$ corresponds to either phonemes or word-pieces; $y_t$ always corresponds to word-pieces), similar to \cite{Tang2022,Zhehuai2022}. 
Note that $x_t$ and $y_t$ are both derived from the same unpaired text.

\begin{figure}[h!]
  \centering
  \includegraphics[scale=0.35]{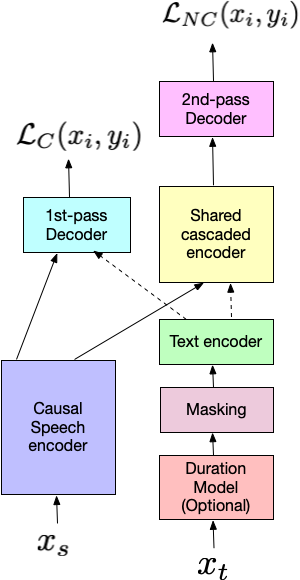}
    \vspace{-0.1in}
  \caption{JOIST architecture for training with speech and text inputs in the cascaded encoder framework~\cite{arun21cascade}.}
  \vspace{-0.1in}
  \label{fig:joint_speech_text}
\end{figure}

Our proposed model, which we call JOIST, is depicted in Figure~\ref{fig:joint_speech_text}. 
The model is based on the cascaded encoder framework~\cite{arun21cascade}, which contains two Hybrid Autoregressive Transducer (HAT)~\cite{Variani20} decoders: the first (blue) operates on the output of the \emph{causal speech encoder} (i.e., zero right context frames); the second (pink) operates on the \emph{shared non-causal cascaded encoder} which has access to 900ms of right context frames.
We refer to these two decoders as the first-pass and the second-pass decoders, respectively.
The cascaded encoder framework is motivated by the goal of having a low output latency decoder (the causal decoder) which can be used to quickly display first-pass results to the screen; the outputs of the non-causal decoder (delayed by 900ms, because of the right context) can be computed in parallel and used to update the results displayed to the screen later following~\cite{sainath2021cascadedlm}.
We denote the output probabilities from the \emph{first-pass} decoder on the paired audio-text as $P_\text{C}(y_s|x_s)$, and the probabilities from the \emph{second-pass decoder} on the paired audio-text as $P_\text{NC}(y_s|x_s)$. 

To be able to train with unpaired text-data, we up-sample the input text representation, $x_t$, following~\cite{Thomas2022,Zhehuai2022}.
However, unlike~\cite{Zhehuai2022}, we use a simpler, parameter-free duration model, as described in Section~\ref{sec:text_duration_modeling}.
The up-sampled text representation is masked (to ensure that the task is sufficiently challenging for the model) and fed to a text encoder.
The output of the text encoder can be fed to the first-pass decoder to generate $P_\text{C}(y_t|x_t)$, or to the second-pass decoder after passing through the shared encoder to generate $P_\text{NC}(y_t|x_t)$.

\subsection{Loss Computation}
The model is trained by jointly optimizing both decoders using audio-text pairs in addition to the unpaired text.
If we denote $\Lm_\text{C}(y, x) = -\log P_\text{C}(y|x)$ and $\Lm_\text{NC}(y, x) = -\log P_\text{NC}(y|x)$ as the negative log likelihood of the causal and the non-causal decoders, respectively, we define the overall loss as: 
\begin{equation}
\begin{split}
    \Lm_\text{CE}&=\lambda_1 [\Lm_\text{C}(y_s, x_s) +  \Lm_\text{NC}(y_s, x_s)] \\
    &+\lambda_2[\Lm_\text{C}(y_t, x_t) +  \Lm_\text{NC}(y_t, x_t)]
    \label{eq:total_loss}
\end{split}
\end{equation}
where, $\lambda_{1}$ is the weight corresponding to the paired audio-text data and $\lambda_2$ is the weight on the unpaired text-only data.
As can be seen in~\eqref{eq:total_loss}, we weight the casual and non-causal decoders equally in the loss function.
In practice, the losses are computed over a mini-batch of examples; in training, we use $50\%$ paired audio-text and unpaired $50\%$ text examples in each mini-batch. 
Unlike previous work, we do not add additional MLM or consistency losses from the text encoder~\cite{Ankur2021, Tang2022, Zhehuai2022} which simplifies the overall training procedure.
Evaluations of the impact of these and other losses in the JOIST framework are left as future work.

\subsection{Duration Modeling to Up-Sample Text Representations \label{sec:text_duration_modeling}}
Previous works~\cite{Thomas2022, Zhehuai2022} have demonstrated the importance of up-sampling the text representations in order to create representations that can be easily aligned with the speech modality.
In this work, we consider a number of schemes for this purpose.

\myparagraph{Fixed Repetition:} In this scheme, each text sub-word unit (word-piece or phoneme) is replicated a fixed number of times, exactly following the approach proposed in~\cite{Thomas2022}.
The drawback of this approach is that this does not match the actual expected durations of various sub-word units in practice (e.g., vowels tend to be longer than consonants; word-pieces with more characters tend to have longer durations).
Thus, repetition by a fixed amount is a simple but crude approximation.
We consider a fixed repetition length of 3, which corresponds to 180ms per unit of $x_t$, in this work.

\myparagraph{Random Repetition:} To address the shortcomings of fixed repetition, we also consider random repetition.
In this approach, the text representation is varied by randomly repeating each unit by sampling from a uniform distribution between 1 and 3 (i.e., 60ms, 120ms, or 180ms per unit of $x_t$).
The potential benefit of random repetition is that it simulates some of the variation that we might expect to see in the distribution of sub-word units.
It must be noted however, that as with the fixed repetition scheme, this is still a crude approximation.

\myparagraph{Sub-Word Distribution}: In this approach, we model the distribution of each sub-word unit using a Gaussian distribution.
Given the paired audio-text data, we generate forced-alignments using a baseline system~\cite{Golan16} to estimate phoneme and word alignments for each word in the transcript; these are used to compute statistics of the number of frames corresponding to each phoneme or word in the supervised training data.
We decompose each word into its constituent word-pieces and evenly distribute the words total frames amongst its constituent word-pieces. 
Thus, by accumulating statistics over the entire training set we can compute the sufficient statistics of the Gaussian distribution -- the mean and standard deviation -- for each unit.
We can repeat each unit by sampling from it's corresponding Gaussian distribution.
This is more exact then fixed or random repetition, but is still an approximation since it ignores contextual effects as each unit is sampled independently.

\myparagraph{Align+Sub-Word Distribution:} We can always use all of the text in the paired audio-text set, $\mathcal{S}$, to augment the unpaired text data, $\mathcal{T}$ -- in effect treating the text in the paired data as unpaired text.
In this specific case we up-sample text examples in $\mathcal{T}$ based on the true number of frames for each unit, obtained using a forced-alignment~\cite{moreno1998recursive}; as before,we divide up the total number of frames in the word amongst its constituent word-pieces.
For text data in $\mathcal{T}$, for which audio (and thus, forced-alignments) are not available, we use subword distribution to up-sample the text.

\subsection{MWER} \label{sec:mwer}
Instead of optimizing model log-likelihoods as in~\eqref{eq:total_loss}, the Minimum Word Error Rate (MWER)~\cite{prabhavalkar2018minimum} strategy minimizes the expected number of word errors.
Specifically, in the standard MWER criterion, given a speech utterance, $x$, corresponding ground-truth text, $y^*$, and a set of N-best hypotheses, $y_i$, ($1 \leq i \leq N$), we minimize the MWER loss proposed in~\cite{prabhavalkar2018minimum}:
\begin{equation}
    \Lm^\text{MWER}(y^*, x) = \sum_{y_i} \left[\frac{P(y_i|x)}{\sum_i P(y_i|x)} \right] \left[\Wm(y_i, y^*) - \frac{\sum_i \Wm(y_i, y^*) }{N} \right]
    \label{eq:mwer}
\end{equation}
where, $\Wm(y, y^*)$ corresponds to the number of word errors between the hypothesis, $y$, and the ground-truth, $y^*$. 
To stabilize training, the MWER loss is interpolated with the standard cross-entropy loss, after initializing from a model that has converged under the maximum likelihood criterion in~\eqref{eq:total_loss}.
MWER training has been shown to improve WER by 5--20\% in previous works~\cite{prabhavalkar2018minimum, GuoTiwariDroppo20, WengYuCui19}.

In the present work, we adapt MWER training by noting that we can compute the MWER loss using the paired audio-text data (i.e., the standard MWER loss paths using $(x_s, y_s)$), but \emph{also using the unpaired text representations} (i.e., $(x_t, y_t)$).
This leads to a novel MWER loss formulation which allows us to train the model using both paired audio-text as well as the unpaired text:
\begin{equation}
\begin{split}
    \Lm &= \lambda_1 \left[ \Lm^\text{MWER}_\text{C}(y_s, x_s) + \Lm^\text{MWER}_\text{NC}(y_s, x_s)\right] \\
    %&+ \lambda_2 \left[\Lm^{MWER}_\text{C}(y_s, x_t) + \Lm^{MWER}_\text{NC}(y_s, x_t)\right] \\
    &+ \lambda_2 \left[\Lm^\text{MWER}_\text{C}(y_t, x_t) + \Lm^\text{MWER}_\text{NC}(y_t, x_t)\right] + \alpha \Lm_\text{CE}
    \label{eq:mwer_st}
\end{split}
\end{equation}
where, $\Lm^\text{MWER}_\text{C}$ and $\Lm^\text{MWER}_\text{NC}$ represent the MWER losses in~\eqref{eq:mwer} computed using the first-pass and second-pass decoders, respectively, and $\alpha$ represents the interpolation weight for the CE loss.
As with CE training in~\eqref{eq:total_loss}, we weight causal and non-causal losses equally.

\subsection{Streaming Metrics \label{sec:latency}}
An important focus of our work is to ensure that the model can be used to produce streaming first-pass recognition results with low latency (i.e., the time between when the user speaks, and the system outputs a sub-word unit). Since we are now injecting an additional source of text-data, where naturally RNN-T would prefer to delay and see more text, it is important to ensure that latency metrics are not degraded with JOIST. We therefore quantify the \emph{streaming quality} of the system in terms of the following latency metrics.

In streaming ASR systems, it is important to detect when the user has finished speaking, so that next fulfillment step can be triggered as quickly as possible. \emph{Endpointer latency} measures the time difference between when the user finishes speaking and when
the system predicts an end of sentence (EOS) token~\cite{bo21system, Shuoyiin19}; a lower endpointer latency allows for faster fulfillment and system response and is thus desirable.
We report both the median (i.e. 50th percentile, \textbf{EP50}) and the 90th percentile (\textbf{EP90}) endpointer latency.

An additional desirable feature is to ensure that the system also has low latency while outputting all intervening words -- i.e., low latency for the partial hypotheses generated by the first-pass decoder, which  will be displayed to the screen.
We measure this by computing \emph{partial latency} -- the time difference between when the first correct partial hypothesis is generated by the model and when the user finishes speaking~\cite{bo21system}.
In this work, we report 50th (\textbf{PR50}) and 90th percentile (\textbf{PR90}) partial latency.

Finally, in order to create the best user experience, we would like to ensure that the hypotheses generated by the first- and second-pass decoders are as similar as possible. 
If not, the outputs presented on the screen will change constantly, which causes \emph{too much screen flickering}.
The \emph{Prefetch Hit Rate} (\textbf{PFHR}) calculates the percentage of utterances where the hypotheses flip between the first- and second-pass decoders, at the utterance level.

\subsection{Novelty of Proposed Method}
Now that we have described JOIST, in this section we further highlight its novelty. First, most previous works which have investigated techniques to directly incorporate unpaired text data into the model have focused on non-streaming encoder-decoder architectures~\cite{YusufGandheSokolov22, Sainath2020b, Ankur2021, Ankur2022, Tang2022, ChungZhuZeng21, AoWangZhou22}.
Although two recent works~\cite{Thomas2022, Zhehuai2022} apply these approachces to an RNN-T model, they only consider full-context (i.e., bi-directional) encoders, and are thus unsuitable for streaming speech recognition.
To the best of our knowledge, our work, JOIST, is the first to investigate whether it is possible to improve \emph{streaming} end-to-end transducer models using unpaired text without synthesis \cite{ZhehuaiTts4Asr2022}.
An additional benefit of the proposed approach is its simplicity: unlike previous works, we focus on joint multi-task training of supervised and unsupervised objectives using a parameter-free duration model, thus greatly simplifying the overall process.
The proposed techniques also demonstrate that it is possible to optimize the model for ASR criteria such as minimum word error rate (MWER)~\cite{prabhavalkar2018minimum} using unpaired text data, which opens up new research directions. 
To the best of our knowledge, our work is the first to demonstrates that it is possible to obtain gains using text even when using large-scale supervised training sets.
\section{Experiments} \label{sec:experiments}
\subsection{Training Sets}
The proposed techniques are evaluated on a large-scale voice search task.
Our first set of experiments are conducted using a supervised training set, referred to as \emph{Train Set A}, that consists of $\sim$300 million United States English multidomain audio-text pairs, which include domains such as Search, Dictation, YouTube and Telephony~\cite{Arun19}.
All domains are anonymized and hand-transcribed, except for YouTube where the transcription is done in a semi-supervised fashion~\cite{liao2013large}. 
Since the effectiveness of various techniques often reduces as the size of the training data increases, in order to test robustness we also consider an even larger dataset, \emph{Train Set B}, which consists of $\sim$650 million United States English multidomain audio-text pairs, spanning similar domains as above; the `supervised' text corresponding to these utterances is obtained using a 600M-parameter teacher system trained on \emph{Train Set A}~\cite{Seong22}.

In addition to the diverse multi-condition training sets, we increase robustness by using multi-condition training data to simulate noisy conditions~\cite{kim2017mtr}; generating data at both 8KHz and 16KHz, with equal probability, to reduce acoustic mismatch due to sampling rates~\cite{YuSeltzerLi13, Li12}; and using SpecAug~\cite{Park2019}.
Noisy data is generated at signal-noise-ratio (SNR) from 0 to 30~dB, with an average SNR of 12~dB, and with T60 times ranging from 0 to 900ms, averaging 500ms.
Noise segments are sampled from YouTube and daily life noisy environmental recordings.

Our unpaired text data consists of more than 100B utterances and spans the domains of Maps, News, Google Play, Web and YouTube, and is thus more than two-orders of magnitude larger than our supervised sets.
In addition, we incorporate all text data from the supervised sets, Train Sets A and B, which we add to the unpaired text data.
In order to ensure that the text data does not degrade quality on the base voice search task, we sample text data from the unsupervised and supervised sets with the same probability so that each unpaired-text minibatch contains 50\% data from Train Set A/B and 50\% unpaired text data, following the standard practice for training N-gram~\cite{Allauzen11} and maximum-entropy~\cite{Biadsy14} LMs. 

\subsection{Evaluation Sets}
Results are reported on multiple test sets which measure the systems ability to recognize the \emph{head} (i.e., relatively common words) as well as the \emph{long tail} of rare words.
The \emph{Search} test set includes around 12K Voice Search utterances with an average length of 5.5 seconds.
They are anonymized and hand-transcribed, and are representative of Google's Voice Search traffic.
In addition, to measure accuracy while recognizing the long-tail of rare words, we create synthetic \emph{rare  word test sets}, as described in~\cite{Peyser2020}. Specifically, we look for words in the LM training data that occur rarely (i.e., less than 5 times) in the supervised training sets A and B.
We construct test sets for each of the 5 domains (i.e., Maps, News, Play, Query, YouTube) by selecting utterances containing rare words and synthesizing them using a TTS system~\cite{Gonzalvo16}.

\subsection{Modeling Architecture} \label{sec:exp_arch}

Our proposed JOIST architecture is modeled as follows. All speech-text pairs use a 128-dimensional log-mel feature frontend computed on 32 msec windows with a 10ms hop.
Features from four consecutive frames are stacked together, and sub-sampled by a factor of 3 to generate 512-dimensional features at a 30ms frame rate. 
These are appended with 16-dimensional one-hot domain-id vectors~\cite{Arun19}, to obtain $x_s$.
The input speech features, $x_s$, are fed to the \emph{causal speech encoder}, which consists of 5 conformer layers~\cite{gulati2020conformer} with 
causal convolution and left-context attention to ensure that the causal speech encoder does not have access to any right context frames.
The self attention layers in the conformer use multi-headed attention with 8 heads, and a convolution kernel size of 15.
We use a stacking layer after the second conformer block, which down-samples the input by a factor of 2, so that the effective frame rate at the output of the causal speech encoder is 60ms.

The text input, $x_t$, is constructed by generating one-hot embeddings of either 4,096 word-pieces ~\cite{Schuster2012} or 46 phonemes depending on the experiment.
These are then up-sampled using the duration model, and masked before feeding them to the text encoder.
Following~\cite{Ankur2021, Joshi2020}, we mask 15\% of the up-sampled text IDs with spans of length 5.
The text encoder is a simple embedding table which takes sub-word IDs as inputs and generates corresponding embeddings.
We set $\lambda_1=0.1$, and $\lambda_2=0.2$, in~\eqref{eq:total_loss} and \eqref{eq:mwer_st}, respectively.

The bulk of the processing of JOIST is performed by the \emph{shared cascaded encoder}, and we devote most of the model's capacity to this block.
The shared cascaded encoder consists of 12 conformer layers; the first 5 have access to three frames of right context each, for a total of $3 \times 5 \times 60\text{ms} = 900\text{ms}$ of acoustic right context.
This specific model structure follows~\cite{Sainath2022}, where it was shown to provide a good trade-off between accuracy and latency.

Both RNN-T decoders (first- and second-pass) consist of a joint network (a single feed-forward layer with 640 units) and an embedding prediction network~\cite{Rami21} which uses an embedding dimension of 640, and conditions on only the last two labels.
In total, each decoder contains 15.2M parameters.
All models use the Hybrid Autoregressive Transducer (HAT) factorization~\cite{Variani20} to predict 4,096 word pieces~\cite{Schuster2012}.  Furthermore, all models are trained with FastEmit \cite{yu21fastemit} to encourage the model to not delay predictions. Overall, the total model size is $\sim$169M parameters.

%\subsubsection{Inference}
For all models, we discard the text-encoder during inference, and evaluate the RNN-T decoders using input speech utterances.
Unless otherwise indicated, all WERs are computed using the second-pass decoder.
Models are decoded with a beam size of 8.

We also compare JOIST to rescoring a lattice generated from the second-pass decoder using a LM.
We train a conformer LM, following~\cite{sainath2021cascadedlm}, which has a look-back attention context of 31.
The LM contains 12 conformer layers~\cite{gulati2020conformer} each of which has a model dimension of 768 and a feed-forward layer dimension of 2048.

\section{Results} \label{sec:results}

\subsection{Full-Context Models}

Since all text injection methods have been explored in the context of full-context encoder layers, in our first set of experiments, we compare previous approaches explored in the literature to our proposed method.
For this set of experiments, we represent the text input representation, $x_t$, using word-pieces, and we use \emph{Train Set A} as our paired audio-text data, $\mathcal{S}$.
Full-context means that the left and right context for self-attention in all conformer layers (both the causal and shared cascaced encoders), is set to allow the model to access all frames in the utterance.
Results of our full context experiments are presented in Table~\ref{table:full_context}, where ``S'' corresponds to the \emph{Search} test set; the rare word test sets are denoted as ``M'' (\emph{Maps}),  ``N'' (\emph{News}), ``P'' (\emph{Google Play}), ``Q'' (\emph{Search Queries}) and  ``Y'' (\emph{YouTube}).

The baseline model, $B0$, is a model that is trained without any text data.
$E0$ corresponds to a JOIST model that uses unpaired text, but does not up-sample the text tokens (i.e., no duration model), similar to~\cite{Tang2022}, except that we do not use a MLM loss and inject wordpieces.
As can be seen in Table~\ref{table:full_context}, without replication, $E0$ is no better then baseline $B0$, and is much worse on the \emph{News} set.
Fixed repetition of the word-piece tokens, which is very similar to~\cite{Thomas2022}, but in a joint training setup ($E1$), however, can achieve a WER improvement of 2--5\% relative over $B0$ on the rare-word test sets.
The alternative duration modeling schemes random distribution ($E2$), and sub-word distribution ($E3$) improve performance over the baseline, $B0$, but do not outperform fixed repetition ($E1$).

Finally, as a comparison to other methods $B1$ shows joint-training of speech-text and text with SLAM \cite{Ankur2021}, where we concatenate the outputs from speech encoder and text encoder before passing to the shared encoder. SLAM works in tasks where the encoder is pre-trained using SLAM, and then fine-tuned using a supervised loss. However, in joint-training the conformer layers learn to compute attention over speech and text jointly in training, which is missing in inference, leading to the high WER.
We compare against MAESTRO~\cite{Zhehuai2022} in the next section.

Our goal in these initial set of experiments confirm the importance of duration modeling and also that joint training is an effective yet simple method to optimize mixed input systems.
We use these initial findings to help guide our experiments with streaming models in the next section, where we also investigate the impact of using phonemes vs. word-pieces as our text representation.
\begin{table}[h!]
    \centering
    \begin{tabular}{|c|c|c|c|c|c|c|c|} \hline
    Exp & Model & S & M & N & P & Q & Y \\ \hline
    B0 & no text & 4.8  & 11.9 & 8.2 & 36.1 & 19.3 & 22.6 \\ \hline
    B1 & SLAM & 75.9  & 87.0 & 99.4 & 87.1 & 84.7 & 91.2 \\ \hline \hline
    E0 & no rep & 4.8 & 11.9 & 8.5 & 35.8 & 19.5 & 22.6 \\ \hline 
    E1 & fixed rep & \textbf{4.6}  & \textbf{11.4} & \textbf{7.9} & \textbf{35.7} & \textbf{18.9} & \textbf{22.1}  \\ \hline
    E2 & random rep & 4.7  & 11.8 & 8.2 & 35.9 & 19.0 & 22.2  \\ \hline
    E3 & sub-wrd dist & 4.9 & 11.8 & 8.0 & 36.2 & 19.5 & 22.2 \\ \hline
    \end{tabular}
    \caption{WER for WPM Text Injection; Full-Context Model}
    \vspace{-0.2in}
    \label{table:full_context}
\end{table}

\subsection{Streaming Models}

First, we repeat the experiment of representing unpaired data $x_t$ in terms of word-pieces, but using a streaming architecture described in Section \ref{sec:exp_arch}.
Once again, we use \emph{Train Set A} as our supervised training set, $\mathcal{S}$.
Our results are presented in Table~\ref{table:streaming_context}. 
Our baseline model ($B2$) corresponds to an E2E model that is trained solely on the paired audio-text data.
We leave out SLAM from the comparison, as Table \ref{table:full_context} showed a degradation due to the speech-text concatenation.
We observe similar trends to the non-streaming model: as long as we up-sample the text representations ($E5$--$E7$), we can obtain a 2--4\% relative improvement in WER over the baseline ($B2$) on the rare word sets; however, JOIST with no repetition $E4$ does not improve over the baseline.
Finally, we also compare to $B3$, a jointly-trained word-piece based MAESTRO model, which uses a TTS-based duration model and consistency losses, with the same cascaded encoder architecture as all other models in the table.
This model is on-par with the parameter-free duration-modeling results from $E5$--$E7$.
Improvements with alternative architectures have been seen with MAESTRO, though we leave that for future work.
\begin{table}[h!]
    \centering
    \begin{tabular}{|c|c|c|c|c|c|c|c|} \hline
    Exp & Model & S & M & N & P & Q & Y \\ \hline
    B2 & no text & 5.2  & 12.5 & 9.0 & 37.4 & 20.0 & 23.2 \\ \hline 
    %B1 & SLAM & 368  & 571 & 92.7 & 779 & 784 & 585 \\ \hline
    B3 & MAESTRO & 5.6  & 12.2 & 9.1 & 36.2 & 20.3 & 23.0 \\ \hline \hline
    E4 & no rep & 5.3 & 12.3 & 15.1 & 36.9 & 20.0 & 23.4 \\ \hline 
    E5 & fixed rep & 5.3  & 12.1 & \textbf{8.8} & 37.0 & \textbf{19.6} & \textbf{23.1}  \\ \hline
    E6 & random rep & \textbf{5.2} & \textbf{12.0} & 9.0 & \textbf{36.7} &	19.8 & 23.3   \\ \hline
    E7 & sub-wrd dst & 5.2 & 12.1 & 9.0 & 36.9 & 19.7 & 23.3 \\ \hline
    \end{tabular}
    \caption{WER for WPM Text Injection; Streaming Model}
    \label{table:streaming_context}
    \vspace{-0.2in}
\end{table}

Next, we contrast the benefits of different duration modeling schemes using phonemes versus word-pieces in representing unpaired text, $x_t$.
These experiments are conducted using the much larger \emph{Train Set B}, which uses a teacher model to generate the paired audio-text data, $\mathcal{S}$.
Our results are presented in Table~\ref{table:phn_vs_wpm}.
We have omitted results on the Search test set for clarity since it typically does not change between techniques, but will present them in the final section.
Since the various word-piece based duration modeling techniques perform similarly, we only list the random repetition baseline $E8$ for brevity.
As can be seen in the table, systems which use phoneme representations ($E9$--$E12$)  outperform word-pieces; There is not a huge difference in performance between the different duration modeling strategies, similar to what we found with word-pieces.
Since $E11$ provides a very slight edge, we will choose that for subsequent exepriments. Overall, $E11$ provides between a 4--14\% relative improvement in WER over $B4$.
The number of phonemes is roughly 3-times that of wordpieces, and it is possible that this finer-granularity is inherently a better model for duration and thus helps with quality.
In the future, we will also compare to a grapheme text representation, to understand if the finer granularity helps, or if gains come because phonemes are typically better for long-tail words compared to graphemes/wordpieces~\cite{Bruguier2019}.

\begin{table}[h!]
    \centering
    \begin{tabular}{|c|c|c|c|c|c|c|} \hline
    Exp & Model & M & N & P & Q & Y \\ \hline
    B4 & Baseline, no Text & 13.9 &	9.4	& 37.9	& 21.6	& 24.4\\ \hline \hline
    E8 & random rep, WPM & 13.7 &	9.3	& 37.6	& 20.9	& 24.2 \\ \hline
    E9 & fixed rep, phn & \textbf{13.0}  &	9.1	& 32.7	& \textbf{18.7} &	21.5 \\ \hline
    E10 & random rep, phn & 13.1 & 9.2 & \textbf{31.2} & 18.9 & 21.3 \\ \hline
    E11 & sub-wrd dst, phn & 13.1 &	9.6 & 32.6 & \textbf{18.7}	& \textbf{21.2} \\ \hline
    E12 & align+dist, phn & 13.1 & \textbf{8.5}	& 33.2	& 19.5	& 22.2 \\ \hline
    \end{tabular}
    \caption{WER for phonemes vs. word-piece based text injection}
    \label{table:phn_vs_wpm}
        \vspace{-0.2in}
\end{table}

\subsection{Comparison to Neural LM}

As stated in the introduction, neural LM is a very common approach used to improve the quality of rare word recognition \cite{Anjuli18}. In this section, we look at a standard cascaded encoder trained only on audio-text pairs ($B4$), that is then rescored by a neural LM ($B5$). The oracle WER of the lattice is also shown in Table \ref{table:neural_lm}.
We compare this to the best phn-JOIST system ($E11$), and also rescore this with a LM ($E13$). Table \ref{table:neural_lm} shows that a LM with the base system ($B5$), still does better then JOIST alone ($E11$) on half of the long-tail sets, though requires an additional 128M parameter LM. However, if we apply the LM to JOIST ($E13$), this outperforms $B5$ on all sets. Moreover, the oracle WER as well as the relative WER improvement of $E13$ is much larger then $B5$, which confirms our hypothesis that JOIST helps to bring rare word hypotheses into the beam, which leads to even better quality with the LM.

\begin{table}[h!]
    \centering
    \begin{tabular}{|c|c|c|c|c|c|c|} \hline
    Exp & Model &  M & N & P & Q & Y \\ \hline
    B4 & no text & 13.9 &	9.4	& 37.9	& 21.6	& 24.4 \\ \hline
    B5 & B4 + neural LM & 10.3 &	9.8 &	33.8 &	14.9 &	20.2 \\
    & oracle &  7.0 & 8.2 & 21.7 & 9.5 & 14.0 \\ \hline
    E11 & JOIST &13.1 &	9.6 & 32.6 & 18.7	& 21.2\\ \hline
    E13 & JOIST + neural LM & \textbf{9.7} &	\textbf{9.4} &	\textbf{28.5} &	\textbf{12.9} &	\textbf{17.1} \\ 
     & oracle &  6.4 & 7.9 & 15.7 & 7.5 & 10.8 \\ \hline
    \end{tabular}
    \caption{WER for Neural LM}
    \label{table:neural_lm}
    \vspace{-0.2in}
\end{table} 

\subsection{Final Best System: Phoneme JOIST}
We take the best system -- the phoneme-based JOIST from $E11$ -- and investigate quality and latency compared to a cascaded encoder model trained only on paired data.

\subsubsection{Quality: MWER Training}
Table \ref{table:mwer} shows $B4$, the baseline cascaded encoder, and $B6$, the baseline after MWER training using paired-only data. In contrast, $E11$ is the JOIST model and $E14$ the model after MWER training, using the method from Section \ref{eq:mwer_st} to train on paired and unpaired data. Further gains are seen with $E14$, particularly on rare-word sets. In addition, we also evaluate the 1st-pass WER for both systems after MWER training, which appears to be around 9.3\%. In the next section, we will discuss metrics around flickering.
\begin{table}[h!]
    \centering
    \begin{tabular}{|c|c|c|c|c|c|c|c|} \hline
    Exp & Model & S & M & N & P & Q & Y \\ \hline
    B4 & CascEnc & 6.2 & 13.9 &	9.4	& 37.9	& 21.6	& 24.4  \\ \hline
    B6 & B4 + MWER & 5.8 & 13.5 & 9.1 & 37.5 & 20.8 & 24.0 \\ \hline \hline
    %E11 &  JOIST & 6.1 &  13.5 &	9.7 & 32.9 & 19.5 & 21.8 \\ \hline
    E11 & JOIST & 6.1 & 13.1 &	9.6 & 32.6 & 18.7	& 21.2 \\ \hline
    %E0 & MWER, s &  \\ \hline 
    E14 & E11 + MWER & \textbf{5.8} & \textbf{12.7} & 9.4 & \textbf{32.0} & \textbf{18.3} &	\textbf{20.7}  \\ \hline
    \end{tabular}
    \caption{WER for MWER Experiments}
        \vspace{-0.2in}
    \label{table:mwer}
\end{table}

\subsubsection{Streaming Metrics}
An important focus of our work is to ensure that JOIST has good streaming recognition performance compared to a cascaded encoder, which we quantify in Table~\ref{table:latency}. First, we see that both the endpointer (EP50, EP90) and partial (PR50, PR90) latencies between JOIST ($E14$) and Cascaded Encoder ($B6$) are on par. Second, the flickering between the 1st and 2nd pass, as measured by PFHR, is also on-par. 
\begin{table}[h!]
    \centering
    \begin{tabular}{|c|c|c|c|c|c|c|c|} \hline
    Exp & EP50 & EP90 & PR50 & PR90 & PFHR\\ \hline
    B6 & 410 & 710 & 20 & 430 & 0.79 \\ \hline
    E14 & 410 & 700 &40	& 430 & 0.79  \\ \hline
    \end{tabular}
    \caption{WER and Latency on Search Test Set}
    \label{table:latency}
        \vspace{-0.2in}
\end{table}

\subsubsection{Comparison on Logs Data}
Finally, we compare the no-text cascaded encoder with phoneme JOIST, by running a ``side-by-side'' (SxS)  on unseen, real-audio search data.
In the SxS, we collect 114 utterances which generate different hypotheses when decoded with the two systems, and send these utterances to two human raters.
Based on these ratings, we report five statistics based on the SxS: \emph{Changed} -- \% of utterances in which the two models produced different hypotheses; \emph{Wins} -- the \# of utts the JOIST is correct and Cascaded Encoder is incorrect;  \emph{Losses} -- the \# of utts JOIST is incorrect and Cascaded Encoder is correct; \emph{Neutral} -- the \# of utts both models are both correct or incorrect; 

The table shows that more than 10\% of the traffic is changed with JOIST, and it has more wins then the Cascaded Encoder. A closer look at the errors shows wins in many rare words, due to the text injection.
\begin{table}[h!]
  \centering
  \begin{tabular}{|c|c|c|c|} \hline
    Changed (\%) & Win & Loss & Neutral \\ \hline
  10.6\% & 23 & 16 & 75 \\ \hline
  \end{tabular}
  \label{table:sxs}
    \caption{SxS: Cascaded Encoder vs. JOIST}
    \vspace{-0.2in}
\end{table}
\section{Acknowledgements}
The authors would like the thank Ruoming Pang, Arun Narayanan, Yanzhang He, Ding Zhao, Shaojin Ding and Fran\c{c}oise Beaufays, for helpful discussions regarding this work.

\bibliographystyle{IEEEbib}

\bibliography{main}
% \bibliography{refNames}
%\input{appendix}
\end{document}